\documentclass[]{article}
\usepackage{proceed2e}
\usepackage{amsmath}
\usepackage{multirow}
\usepackage{array}
\usepackage{booktabs}
\usepackage{graphicx}
\usepackage{pdfpages}
\usepackage{subfig}
\usepackage{float}
\usepackage[linesnumbered,ruled]{algorithm2e}
\usepackage[square]{natbib}
\usepackage{times,txfonts}

\newcommand*\samethanks[1][\value{footnote}]{\footnotemark[#1]}

\title{Tractable Querying and Learning in Hybrid Domains \\ via Sum-Product Networks}

\author{Submission} 
\author{ {\bf Andreas Bueff\,\thanks{~~Contributed equally to the research.}} \\
University of Edinburgh\\
\And
{\bf Stefanie Speichert\samethanks}  \\
University of Edinburgh\\
\And
{\bf Vaishak Belle}   \\
University of Edinburgh \& \\ Alan Turing Institute \\
}

\begin{document}

\newcommand{\ourlearn}{LearnWMISPN}

\maketitle

\begin{abstract}
\small 
Probabilistic representations, such as Bayesian and Markov networks, are fundamental to much of statistical machine learning. 
Thus, learning probabilistic representations directly from data is a deep challenge, the main computational bottleneck being inference that is intractable. Tractable learning is a powerful new paradigm that attempts to learn distributions that support efficient probabilistic querying. By leveraging local structure, representations such as sum-product networks (SPNs) can capture high tree-width models with many hidden layers, essentially a deep architecture, while still admitting a range of probabilistic queries to be computable in time polynomial in the network size. The leaf nodes in SPNs, from which more intricate mixtures are formed, are tractable univariate distributions, and so the literature has focused on Bernoulli and Gaussian random variables. This is clearly a restriction for handling mixed discrete-continuous data, especially if the continuous features are generated from non-parametric and non-Gaussian distribution families. In this work, we present a framework that systematically integrates SPN structure learning with weighted model integration, a recently introduced computational abstraction for performing inference in hybrid domains, by means of piecewise polynomial approximations of density functions of arbitrary shape. Our framework is instantiated by exploiting the notion of propositional abstractions, thus minimally interfering with the SPN structure learning module, and supports a powerful query interface for conditioning on interval constraints. Our empirical results show that our approach is effective, and allows a study of the trade off between the granularity of the learned model and its predictive power. 

\end{abstract}

\section{INTRODUCTION}

Probabilistic representations, such as Bayesian and Markov networks, are fundamental to much of statistical machine learning. Thus, learning probabilistic representations directly from data is a deep challenge, the main computational bottleneck being inference that is intractable (\#-P hard \citep{valiant1979complexity,DBLP:journals/jair/BacchusDP09}). In practise, most systems approximate inference during the learning step as well as the querying step, often with weak guarantees, which is problematic for applications such as autonomous vehicles and health monitoring.

Tractable learning is a powerful new paradigm that attempts to learn distributions that support efficient probabilistic querying. Much of the initial work focused on low tree-width models \citep{bach2002thin}, but later, building on properties such as local structure \citep{chavira2008probabilistic}, data structures such as arithmetic circuits (ACs) emerged. These can also represent high tree-width models and enable exact inference for a range of queries in time polynomial in the circuit size. Sum-product networks (SPNs) \citep{poon2011sum} are instances of ACs with an elegant recursive structure -- essentially, an SPN is a weighted sum of products of SPNs, and the base case is a leaf node denoting a tractable probability distribution (e.g., a univariate Bernoulli  distribution). In so much as deep learning models can be understood as graphical models with multiple hidden variables, SPNs can be seen as a tractable deep architecture. Of course, learning the architecture of standard deep models is very challenging \citep{bengio2009learning}, and in contrast, SPNs, by their very design, offer a tractable structure learning paradigm. While it is possible to specify SPNs by hand, weight learning is additionally required to obtain a probability distribution, but also the specification of SPN has to obey conditions of completeness and decomposability, all of which makes structure learning an obvious choice. Since SPNs were introduced, a number of structure learning frameworks have been developed for those and related data structures, e.g., \citep{gens2013learning,hsu2017online,liang2017learning}. 

Since the base case for SPNs is a tractable distribution, SPNs are essentially a schema for composing   tractable distributions. Although much of the literature focuses on univariate Bernoulli distributions \citep{gens2013learning,liang2017learning}, continuous models can be considered too, which makes SPNs very appealing because real-world data is often hybrid, with discrete, categorical and continuous entities. Nonetheless, this rests on the assumption of having a tractable query interface to the underlying continuous and hybrid distributions. In \citep{hsu2017online}, for example, the leaf nodes are assumed to be drawn from a Gaussian, and similarly, \citep{vergari2015simplifying,molina2017poisson} assume the data is drawn from other families with known parametric form. 

In the literature on probabilistic inference, to handle non-parametric families, in (say) non-Gaussian graphical models, there has been considerable interest in so-called piecewise polynomial approximations \citep{sanner2012symbolic,shenoy2011extended,belle2015probabilistic}. Essentially, these approximations can be made arbitrarily close to (say) exponential families and support efficient integration (e.g., \citep{baldoni2014user,albarghouthi2017quantifying}). Moreover, as shown in \citep{belle2015probabilistic}, just as inference in discrete graphical models reduces to weighted model counting \citep{chavira2008probabilistic,DBLP:journals/jair/BacchusDP09}, obtained from the sum of products of atoms' weights in a propositional model, inference in hybrid graphical models reduces to the so-called weighted model integration (WMI), obtained by integrating the product of atoms' densities in a first-order model. For example, a Gaussian distribution $\mathcal{N}(5,1)$ for a random variable $x$ can be represented using (truncated) models: 

\begin{figure}[h]
  \centering
  \subfloat[Constant Approximation]{\includegraphics[width=4cm]{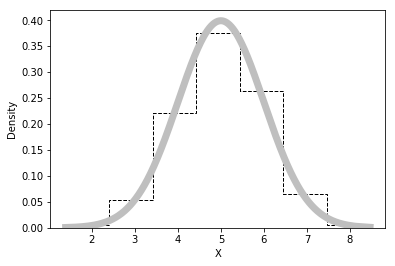}\label{fig:f1}}
  \hfill
  \subfloat[Third Order Polynomial ]{\includegraphics[width=4cm]{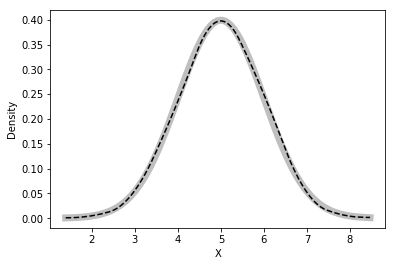}\label{fig:f2}}
  \caption{Comparison of piecewise constant vs polynomial using a 7 piece function}
  \label{fig:comparison}
\end{figure}

As a computational strategy, what makes WMI particularly effective is that: (a) the intervals can be understood as (and mapped to) propositions \citep{belle2015probabilistic,morettin2017efficient}, known in the literature as {\it propositional abstraction}, allowing us to piggyback on any propositional solver, and (b) WMI admits complex interval queries, such as $\Pr( x > 1.5 \mid y <2)$ where $y$ is some other random variable, yielding a powerful query interface. Notice here that the intervals in the queries do not have to correspond to the intervals used for specifying the distribution, and can overlap and subsume them arbitrarily. 

\begin{figure}[h]
  \centering
\includegraphics[width=8cm]{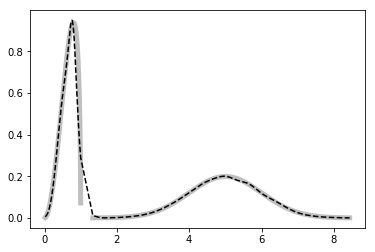}
  \caption{The piecewise polynomial approximate of a mixture of a Gaussian and Beta distribution}
  \label{fig:mixPDF}
\end{figure}

In this paper, we combine the power and flexibility of WMI with SPNs to address the problem of tractable learning and querying in mixed discrete-continuous domains. Among other things, this allows us to learn both parametric and non-parametric models and their mixtures, including models such as Figure \ref{fig:comparison} and Figure \ref{fig:mixPDF}, but with no prior knowledge of the true distribution, all of which is further composed over hidden layers in a deep architecture. In other words, unlike many conventional deep learning frameworks, leveraging SPNs enables a fully   \textit{unsupervised} learning regime for hybrid data. 

From a representational viewpoint, our framework allows the modeller to study the granularity of the continuous distributions (in terms of the number of piecewise components and the degree of the polynomial fit), and determine the empirical trade off between accuracy and model size / interpretability. By lifting the propositional abstraction strategy of WMI, the framework is also instantiated in a manner that allows us to use any SPN learner in principle. To the best of our knowledge, this is a first attempt to combine SPNs and WMI in their full generality, with an interface for complex interval  queries. 

Outside of the above lines of research, research in learning in hybrid domains has been primarily limited to well-known parametric families. For example, \citep{HGC95} focus on conditional linear Gaussian models, and \citep{yang2014exponential} focus on mixing exponential families. This is perhaps not surprising, because inference in hybrid domains has been primarily limited to approximate computations, e.g., \citep{murphy1999variational}, and/or Gaussian models \citep{MixtureofGaussians}. In similar spirit, approaches such as \citep{nitti2016learning,ravkic2015learning} consider learning of complex symbolic constraints by assuming base distributions as being  either Gaussian or a softmax equality. 

We are organised as follows. We introduce the formal foundations for our work, and then turn to our approach that integrates SPNs and WMI. We then discuss our interface for complex interval queries. We turn to empirical evaluations after that, and finally conclude.

\section{Preliminaries} 

\subsection{Weighted Model Integration}

Given a propositional formula $\Delta$, the task of model counting is to compute the total number of satisfying assignments for $\Delta$.  
For example, $p\lor q$ has 3 models, and a satisfying ratio of $3/4$. Weighted model counting (WMC) is its extension that additionally accords weights to the models \citep{chavira2008probabilistic}. Models are usually accorded weights by computing the product of weighted literals: that is, assuming $w$ maps literals in $\Delta$ to positive reals, we define:
\[ \mbox{WMC}(\Delta,w) = \sum_{M\models \Delta} \prod_{l \in M} w(l) \] where the sum ranges over propositional models, and $l\in M$ denotes the literals true at the model $M$. For example, suppose $p$ and $q$ are accorded a weight of .6 and .3 respectively, with the understanding that the weights of a negated atom $\neg a = 1 - w(a)$. Then, the weight accorded to $[p=1,q=1]$ would be $.18$, and $\mbox{WMC}(p\lor q, w) = .18 + .42 + .12= .72$. 

WMC is a state-of-the-art exact inference scheme, and owing to its generality, inference in a number of formalisms, such as relational Bayesian networks and probabilistic programs \citep{DBLP:conf/uai/FierensBTGR11}, are encoded as WMC tasks.  Given a formula $\Delta$, a weight function $w$, a query $q$ and evidence $e$,  probabilities are computed by means of the expression: $\Pr(q\mid e, \Delta) = \mbox{WMC}(\Delta \land q \land e, w)/\mbox{WMC}(\Delta \land e,w).$

Due to the propositional setting, however, WMC is limited to finite domain discrete random variables, which has spurred considerable interest in generalising WMC to continuous and hybrid distributions \citep{belle2015probabilistic,RupakSMT,albarghouthi2017quantifying}. 
Weighted model integration (WMI) is a computational abstraction for computing probabilities with continuous and mixed discrete-continuous distributions \citep{belle2015probabilistic}.  
The key idea is to additionally consider inequality atoms, such as $0\leq x \leq 10$, along with possibly non-numeric weights, such as $x^2$. The intuition is to let the inequality atom denote an interval of the domain of a density function, and let the weight define the function for that interval. Given such a knowledge base, the weighted model integration (WMI) is obtained from the model count of the theory together with a volume computation for the intervals. Formally:
\[ \mbox{WMI}(\Delta, w) = \sum_{M \models \Delta^-} \int\displaylimits_{\{l^+\colon l\in M\}} \prod_{l \in M} w(l) \] where $\phi^-$ denotes a {\it propositional abstraction}, based on a mapping from linear arithmetic expressions to propositional atoms, and $\phi^+$ denotes a {\it refinement}, where the atoms are mapped back to to their linear arithmetic expressions. For example, $(0 \leq x \leq 10) 
\lor q$ on abstraction yields $p \lor q$, where $p$ is a fresh atom chosen so as to not conflict with any of the existing atoms in the theory, which has a model count of 3 on abstraction. Suppose we have weights of $x^2$ and $.3$ respectively, and suppose the weights of all negated atoms are 0. Then, 
the WMI for the model $[p=1,q=1] = [(0\leq x\leq 10) = 1, q=1]$ is obtained as $\int_0 ^{10} x^2 \cdot .3 = 300/3 = 100$. Because negated atoms obtain a weight of 0, the WMI for $p \lor q$ is also 100. Conditional probabilities are calculated using precisely the same expression as for WMC. 

The WMI apparatus is very flexible \citep{morettin2017efficient}: for example, it is possible for query and evidence atoms to arbitrarily range over the intervals mentioned in the knowledge base. 
For example, suppose we have the query atom is $5 \leq x \leq 7$ and $e = {\it true}$.  (We treat the weights of query atoms and their negations as being 1.) In that case, $\Pr(q \mid e, \Delta) = 21.8.$ 

\subsection{Sum-Product Networks} 

SPNs are rooted acyclic graphs whose internal nodes are sums and products, and leaf nodes are tractable distributions, such as Bernoulli and Gaussian distributions \citep{poon2011sum,gens2013learning}. More precisely, letting the {\it scope} of an SPN be the set of variables appearing in it, a recursive definition is given as follows: 

(1) a tractable univariate distribution is an SPN (the base case);
(2) a product of SPNs with disjoint scopes is an SPN (denoting a factorisation over independent distributions); and 
(3)  a weighted sum of SPNs with the same scope, where the weights are positive, is
an SPN (denoting a mixture of distributions). 

Formally, SPNs encode a function for every node that maps random variables $X_1, \ldots, X_n$ to a (numeric) output. For node $k$, the corresponding function $f_k$ maps a world $X_1 = x_1, \ldots, X_n = x_n$ to its probability if its a leaf node, the weighted sum $\sum_i w_i f_i^k (x_1, \ldots, x_n)$ for sum nodes where $f^k_i$ are the children of node $k$, and $\prod_i f_i^k (x_1, \ldots, x_n)$ for product nodes. Conditional probabilities are expressed using: 
\[ \Pr(x_1,\ldots,x_m \mid x_{m+1}, \ldots, x_n) = f_0 (x_1, \ldots, x_n) / f_0 (x_{m+1}, \ldots, x_n)\]  where $0$ is the root node, and the notation $f_k$ with only a subset of the arguments used in the denominator denotes marginalisation over the remaining arguments. Similar expressions can be given for other queries, such as the MAP state and marginals. The benefit of SPNs here is that a bottom-up pass allows us to compute these classes of probabilistic queries in time polynomial in the circuit size. (See, for example, \citep{bekker2015tractable} on other data structures that allow a more expressive class of queries.)

\section{Structure Learning}
We propose an algorithm to derive the structure and parameters for SPNs with piecewise polynomial leaves, equipped with WMI (WMISPNs).  The method \ourlearn\ is capable of deriving such a SPN structure from hybrid data and further, it allows for the retention of the distribution parameters in the continuous case for use in advanced querying. 

The structure learning approach for \ourlearn\ is derived from LearnSPN \citep{gens2013learning}. LearnSPN is a recursive top-down learning method which is capable of learning the structure and weights for SPNs by identifying mutually independent variables, and clustering similar instances. LearnSPN takes a simple approach to structure learning by recursively splitting data into a product of SPNs over independent variables, or a sum of SPNs comprising subsets of instances. More generally, LearnSPN represents an algorithm schema which allows for a modular approach in structure learning from differing domains, which is why our implementation changes only the base case.

\begin{figure}[h]

\includegraphics[width=0.5\textwidth]{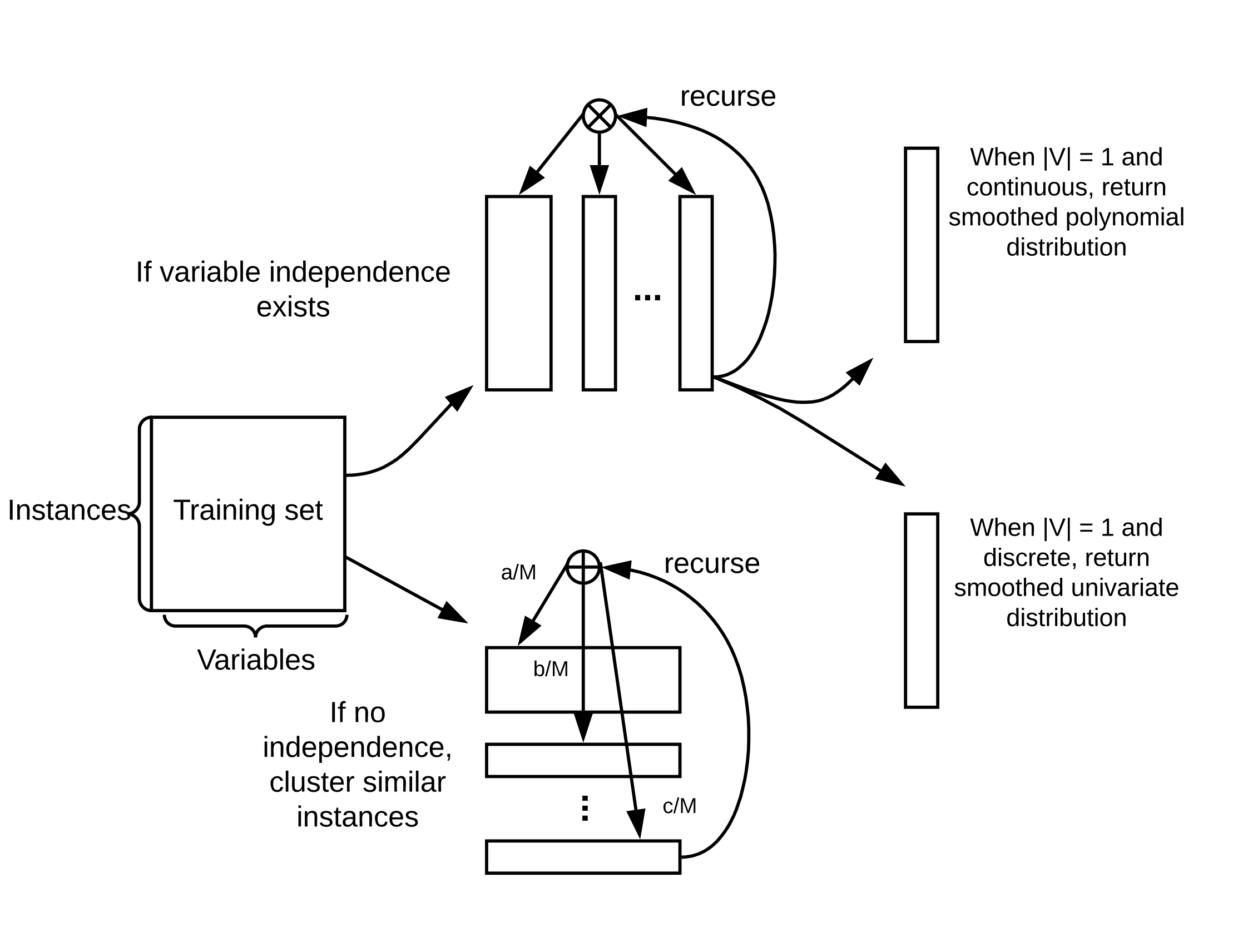}
\caption{A recursive algorithm for learning hybrid domains}
\label{fig:algpic}
\end{figure}

The primary extension with \ourlearn\ is the addition of generated polynomial leaf nodes. 
Given a dataset with discrete, categorical and continuous attributes, 
\ourlearn\ learns  polynomial weight functions as density approximations for the continuous cases, mapped to leaf nodes. By doing so, \ourlearn\  provides a model which is capable of complex querying in the continuous case, while also conditioning over other variable types for hybrid domains.

Prior to structure learning, a preprocessing step is performed which first transforms discrete, categorical, and continuous features into a  binary representation. Much like the algorithm schema of LearnSPN, the preprocessing step can implement any number of unsupervised binning methods including using a \textit{mean split, equal frequency binning,} or \textit{equal width binning}. This is followed by a polynomial learning function which identifies the polynomial function coefficients and probability densities for the continuous features in the hybrid domain  without prior knowledge of the true density function. 

\begin{algorithm}
    \SetKwInOut{Input}{Input}
    \SetKwInOut{Output}{Output}

    \Input{ $T$: set of instances, $V'$: set of variables, $ \theta$: parameters (cf. Algorithm 2) }
    \Output{SPN representing a distribution over $V'$ learned from $T$}
    \eIf{$|V'|= 1$}
      {
        \eIf{\textnormal{variable} $V'$ \textnormal{is an instance of a continuous feature}}{return univariate\ distribution\ 	estimated\ \newline from\ polynomial\ probability\ densities}{return univariate\ distribution\ estimated\ \newline from\ binary\ counts\ in\ $T$}
        		           
      }
      	{partition\ $V'$\ into\ approximately\ independent\ subsets\ $V'_j$      
        \eIf{\textnormal{success}}
      {
        return $\prod_j$\ourlearn$(T,V'_j)$\;
      }{partition\ $T$\ into\ subsets\ of\ similar\ instances\ $T_i$ \newline return $\sum_i \frac{|T_i|}{|T|} $\ourlearn$(T_i,V')$}
   }
    \caption{\ourlearn\ $(T, V')$ }
    \label{alg:1}
\end{algorithm}

Algorithm \ref{alg:1} describes the recursive structure learning algorithm for \ourlearn\ and Figure \ref{fig:algpic} illustrates the recursive pipeline. \ourlearn\ takes as input the preprocessed binary dataset $D'$ with $T$ instances and $V'$ variables and returns a WMISPN representing the distributions of the hybrid domains. \ourlearn\ recurses on subsets of $V'$ until a vector of unit length is found, at which point a corresponding univariate distribution is returned. 

The main novelty in \ourlearn\ is the handling of the base case. Without any prior knowledge, \ourlearn\ will return a smoothed univariate distribution for discrete and categorical variables, or piecewise polynomial distribution for continuous variables. These two outcomes sets \ourlearn\ apart from LearnSPN, and this is represented with polynomial leaf nodes corresponding to continuous distributions. 

If the base case has not been reached, then \ourlearn\ continues the recursive structure learning of LearnSPN. For the decomposition step, a variable split is identified which results in mutually independent subsets, generating a product node for the resulting subset. Mutual independence for the variable splits is determined by a G-test for pairwise independence. For the conditioning step similar instances are clustered using hard incremental expected-maximization (EM) generating a sum node. The ratio split on the  clustered subset of instances determines the corresponding weight and returned is the sum of the resulting weighted clusters.

\paragraph{Preprocessing Step:}
A preprocessing step is required to first transform the discrete, categorical, and continuous variables into a  binary representation. The methods for converting a dataset into a binary representation is contingent on the variable type in the hybrid dataset. The preprocessing pipeline is summarized in Algorithm \ref{alg:2}.

\begin{algorithm}
    \SetKwInOut{Input}{Input}
    \SetKwInOut{Output}{Output}

    \Input{ $D$: dataset of real values, $\textbf{f}$: vector of domain types, $b$: number of bins }
    \Output{$D'$: binary representation of $D$, $M$: matrix of coefficients for the polynomials, $\textbf{p}$:vector of probability densities}
    $\qquad D' \leftarrow$EqualWidthBinning$(D,\textbf{f}, b)$ 
    \newline
       $\qquad M,\ \textbf{p} \leftarrow$PolynomialLearner$(D,\textbf{f}, b)$ 
    \caption{Preprocessing step$(D,\textbf{f},b)$ }
    \label{alg:2}
    
\end{algorithm}

The first sub-task converts the real valued dataset $D$ into a binary representation $D'$. In our model, equal width binning was performed on continuous features, with the number of equal width bins specified as a parameter. (A meta-learning step can be designed to choose the optimal number by studying log-likelihood or polynomial improvements.)  
Higher bin counts would increase the representative power and improve accuracy for advanced queries on continuous distributions (see section 5). Each new bin generated from a feature represents a discrete range on that continuous distribution. A one-hot encoding transformation was performed on the expanded feature set to get a binary representation. For discrete and categorical variables, the binary representation is achieved again with a one-hot encoded transformation with equal width binning not taken into consideration.

As an example for one-hot encoding in the discrete case, consider a random variable $Y$ which has a domain of boolean values. Booleans result in two classification labels so $Y$ is expanded to $(Y_1, Y_2)$, which now correspond to each boolean instance. Formally:
\[
Y(x) = \begin{cases}
               Y_1 = 1\ \mbox{if}\ x  \\
               Y_2 = 1\ \mbox{if}\ \neg x \\
   
            \end{cases}
\]
The two cases for $Y$ are then represented as $[x] \rightarrow [1, 0]$ and $[\neg x] \rightarrow [0, 1]$ respectively. Now consider the continuous case, where equal width binning is implemented, we say that random variable $X$ is defined as having a range $[a,b] \implies {x \in \mathbb{R}:\ a \leq x \leq b}$. If we split $X$ into say three bins $(X_1,X_2,X_3)$, with the split point defined as $s = (b - a)/3$ then we again have another representation for one hot encoding. Formally:
\[
X(x) = \begin{cases}
               X_1 = 1\ \mbox{if}\ a\leq x < (a+s) \\
               X_2 = 1\ \mbox{if}\  (a+s) \leq x < (a + 2s) \ \\
   X_3 = 1\ \mbox{if}\  (a+2s) \leq x\ \\
            \end{cases}
\]

So for the minimum and maximum values in $X$ we get the following one-hot encoded representation $[a]\rightarrow[1,0,0]$ and $[b]\rightarrow[0,0,1]$. This means the binary transformation of an instance $D_{i,:} = [b,x]$ is $D'_{i,:} = [0,0,1,1,0]$. 

\paragraph{Polynomial Learner:}The second task learns a piecewise polynomial approximation for any univariate probability density function (PDF) without prior knowledge or simplifying assumption. 
The original  dataset $D$, the indices for the continuous variables, and the number of bins are taken as input, and as output we receive a vector of probability densities for each continuous feature bin. 
This bin information is retained in its respective polynomial leaf node in the SPN for use in advance querying. 

The piecewise polynomial functions are calculated through a linear combination of b-splines \citep{Speichert:Thesis:2017}. Basis splines form a basis in the piecewise polynomial space. Linear combinations of splines can represent any piecewise polynomial function. This is calculated while ensuring that the piecewise polynomial follows the usual properties of density functions such as its integral being 1.

The order, and, if not previously specified, the number of bins are chosen during training time by means of the Bayesian Information Criterion (BIC) \citep{schwarz1978estimating}. This scoring function has been shown to be robust and was chosen to avoid overfitting the model parameters.

The procedure is performed without any knowledge of the true underlying density function. It, furthermore, displays some desirable properties for tractable inference such as polynomials being computationally easy to find and integrate and being closed under multiplication and addition and, therefore, under mixtures and marginalisation. In addition, the chosen method favours smaller polynomial ranks \citep{Speichert:Thesis:2017} which improves the efficiency of the integration without loosing accuracy.

\section{Querying}
SPNs provide tractable querying on univariate distributions, and have the capacity to calculate the marginal and conditional probabilities on learned features. Other schemas for SPNs limit the scope of querying to binary activations  Pr$(X_i = 1) = x_i$  to the leaf nodes \citep{poon2011sum}. In WMISPNs, we are able to expand our queries to complex cases where $f_k$ maps $X_i$ to new probability ranges $(a \leq x \leq b)$, $a,b \in \mathbb{R}$.
\subsection{SPN Inference}
General SPN inference is initiated at the leaf nodes where queries are presented in the form of selected activations of the random variables $Q = [X_1=1,X_2 = 0, \dots, X_n = 1]$. The mapped probabilities $x_i$ propagate upward from the leaf nodes where values from children nodes are either multiplied at product nodes, or the weighted sum is taken at sum nodes. The final likelihood of the queries is returned at the root node $0$.  

In order to normalize the returned likelihood, a partition function is required to calculate the normalization constant. Formally: \[ Z = \sum_i f_0(x_i)  \] Inference is deemed tractable due to the partition function being computational in time linear to the number of edges in the learned SPN, which in turn results in queries that can be calculated in time linear Pr$(Q) = {f_0(x_1,1 - x_2,\dots,x_n)}/{Z}$. The same is true for MAP and conditional likelihood estimates.

\subsection{Complex Querying}

We extend the SPN inference model to allow complex queries where inference can be performed over continuous as well as discrete features, and combinations thereof: for example, queries on discrete features such as $male(X)$ conditioned on continuous ranges such as $40 \leq weight(X) \leq 50$.

The probability of a query is calculated in the relevant leaf node(s). In the previous section, we introduced an algorithm that assigned ranges $r_i =[\alpha_i,\beta_i]$ and their piecewise polynomial density  approximation $p_i(x)$ to a continuous SPN leaf node $i$. With that, we can perform inference at the leaf node itself through WMI. For example, assume a continuous leaf node for the attribute ``weight" has the range $34 \leq weight(X) \leq 55$ and suppose its calculated polynomial is $-0.051 + 0.0016x$. The probability for the query $40\leq weight(X) \leq 50$ can then be calculated using  $\int_{40}^{50} -0.051 + 0.0016x \quad dx =0.21$.

Once the leaf has processed the query its value is passed to the SPN where it is applied to the other variables of the query. We reiterate that  inference using the polynomials is not performed in the SPN itself. Only the value is passed on to be interpreted. This demonstrates the built-in modularity of our approach.

Naturally, posing a query such as $(40\leq weight(X) \leq 60)$, spanning over multiple intervals   is less trivial. In this case, assume a second interval for $weight(X)$ is specified $55\leq weight(X) \leq 77$ with a polynomial density of $0.1469 - 0.0019x$. Then, the probability of the query is  calculated as  $P(40 \leq weight(X) \leq 70)=P(40 \leq weight(X) \leq 55) + P(55 \leq weight(X) \leq 70) = \int_{40}^{55} -0.051 + 0.0016x \quad dx + \int_{55}^{60} 0.1469 - 0.0019x \quad dx = 0.375+0.291=0.666$.
In general, as the ranges were defined to be mutually exclusive, the calculation of the probability boils down to:
\begin{enumerate}
\item dividing the range into $m-k$ subranges where each corresponds to a leaf node $i, \quad i \in \{1,...,n\}$, $1\leq k \leq m \leq n$; 
\item performing inference through WMI in each leaf with the polynomial density approximation $p_i(x)$; and then 
\item adding the calculated probability values to obtain the final value.
\end{enumerate}

That is, 
\[\int_a ^ b p(x) dx = \int_a^{\beta_k} p_k(x) dx + \sum_{j=k+1}^{m-1} \int_{\alpha_j}^{\beta_j} p_j(x) +  \int_{\alpha_m}^b p_m(x) dx 
\]

By extension, if a query consists of multiple continuous features $X_1,...,X_N$, with density approximations $p^1(X_1),...,p^N(X_n)$, the probability can be calculated as:
\[ \int_{a_1}^{b_1} ... \int_{a_N}^{b_N} p^1(x_1) ... p^N(x_N) dx_1... dx_N
\]
\[= \int_{a_1}^{b_1} p^1(x_1) dx_1 \times ... \times \int_{a_N}^{b_N} p^N(x_N) dx_N.
\]
When integrating the model into the SPN some SPN properties need to be considered. The root node will always calculate the likelihood of a query by taking the product of its children. Given the mutual exclusivity of a continuous feature's multiple ranges, two or more passes are required for an SPN to calculate the likelihood of a query atom over multiple intervals. 
In addition, queries containing multiple independent variables including combinations of continuous feature intervals and discrete query atoms, are passed separately and the product of the returned likelihoods is calculated to obtain the final value.

For simplicity, we assume that queries are limited to the following syntax:

\begin{enumerate}
\item Every query atom is either an interval $x \in [a,b]$ or a discrete feature A $\in [a_1,...,a_n]$.
\item A query consists of a number of conjunctions. 
\item Each conjunction has to be distinct, e.g. no conjunction shares the same query atoms. 
\end{enumerate}

Here, (2) and (3) are not fundamental restrictions:  it is easy to extend (2) by applying the rule $\Pr(A \lor B) = \Pr(A) + \Pr(B) - \Pr(A \land B)$, and in the case of (3), linear arithmetic solvers can be used to reason about multiple constraints for the same variable: for example, $\Pr(30\leq weight(X)) \land \Pr(50\leq weight(X) < 60)$ can be resolved to $\Pr(50\leq weight(X) < 60).$ So, (1)  says computing the probability of expressions such as $x>y$ and $x + y > 2z$, where $x,y,z$ are continuous variables,  cardinality queries \citep{bekker2015tractable}, among others are not dealt with currently.

\section{Experimental Evaluation}

The goal of this section is to evaluate the merits of \ourlearn, and the complex query interface. Specifically, we attempt to address the following questions: 

\noindent \textbf{Q1}~~How effective (in terms of log-likelihood) is \ourlearn\ on complex mixed discrete-continuous data?

\noindent \textbf{Q2}~~How reasonable is the learned distribution: that is, what is the order of the learned polynomials, and what is the spread of the probabilities for the underlying intervals? 

\noindent \textbf{Q3}~~How effective is the query interface in the presence of increasing query lengths? 

We measured the performance of the learned SPNs on preprocessed hybrid domain datasets discussed in an independent and very recent effort \citep{molina2018mixed} that introduces so-called mixed SPNs (MSPNs). Like our work, they are motivated by piecewise approximations for learning from hybrid data. However, we have strived for a full integration of SPNs and WMI. For example, we learn polynomials of the optimal order (by using BIC), whereas they only focus on piecewise constant or linear. We dynamically handle arbitrarily complex interval queries; no such interface is studied there. Perhaps most significantly, our approach neatly separates the propositional layer from the continuous aspects, especially the integration, which makes our framework generic, and applicable to any SPN learner. In particular, we are able to piggyback on LearnSPN's simple yet fast decomposition scheme.

Our performance evaluation of SPNs in conjunction with WMI was performed using three different regimes. We measured the performance of the learned SPNs from \ourlearn\ on preprocessed hybrid domain datasets. Doing so allowed us to directly compare the performance of \ourlearn\ to LearnMSPN \citep{molina2018mixed}, and test whether our approach with continuous distributions was effective. We then investigated whether further increasing the number of leafs per continuous feature resulted in improvements in the model. 
Next, measuring the complex querying capacity of WMI with SPNs was performed by recording the computation times dependent on query length. As we mentioned before, WMI is a very flexible framework for inference in hybrid domains, and as our empirical results show, our integration with SPNs has yielded a \textit{scalable} unsupervised learning architecture that is able to handle MSPN benchmarks and many (preprocessed) UCI datasets involving hundreds of variables. 

\paragraph{Q1 Hybrid Benchmarks:}

We evaluated \ourlearn\ on 11 hybrid domain datasets taken from the UCI machine learning repository \citep{Dua:2017}. Each dataset was composed of differing proportions of categorical, discrete, and continuous features. The diverse set of domains comprised data from financial, medical, automotive and other sectors. MSPNs were used as a comparative baseline. MSPNs proved to be a successful mixed probabilistic model \citep{molina2018mixed}, and given the similar nature of our objectives, a comparison was appropriate. Structure learning with LearnMSPN was performed with different variants. MSPNs were trained using the Gower distance, which is a metric over hybrid domains, with Gaussian distributional assumptions for continuous variables. MSPNs were also trained using the randomized dependency coefficient (RDC) which does not make parametric assumptions. For each model, histogram representations were compared against isometric regression models. 

The dimensions of the datasets can be seen in Table \ref{datastats-table}. Given the original datasets used real values, our preprocessing results in an augmented binary matrix that has a larger variable count. In the original MSPN work, more datasets were used to measure performance, but in our case we omitted datasets which did not contain continuous features. We used the original LearnSPN training, validation, and test ratio splits, which were 75\%, 10\%, and 15\% respectively.

\begin{table}[h]
\caption{Dataset statistics}
\label{datastats-table}
\begin{center}
\begin{tabular}{|p{1.2cm} |p{0.7cm}| p{0.7cm}| p{0.7cm} |p{0.7cm} |p{0.7cm}| }
\hline
\bf Dataset & $|V|$ & $|V'|$ & Train & Valid & Test\\
\hline
\scriptsize anneal-U & \scriptsize 38& \scriptsize 95 & \scriptsize 673 & \scriptsize 90& \scriptsize 134 \\

\scriptsize australian & \scriptsize 15 & \scriptsize 50 & \scriptsize 517 & \scriptsize 69 & \scriptsize103  \\

\scriptsize auto & \scriptsize 26 & \scriptsize 85 & \scriptsize 119 & \scriptsize 16 & \scriptsize23  \\

\scriptsize car & \scriptsize 9 & \scriptsize 50 & \scriptsize 294 & \scriptsize  39& \scriptsize 58 \\

\scriptsize cleave & \scriptsize 14 & \scriptsize 35 & \scriptsize 222 & \scriptsize 29 & \scriptsize44  \\

\scriptsize crx & \scriptsize 15 & \scriptsize 54 & \scriptsize 488 & \scriptsize 65 & \scriptsize 97  \\

\scriptsize diabetes & \scriptsize 9 & \scriptsize 33 & \scriptsize 576 & \scriptsize  76& \scriptsize115  \\

\scriptsize german & \scriptsize 21 & \scriptsize 76 & \scriptsize 750 & \scriptsize 99 & \scriptsize150  \\

\scriptsize german-org & \scriptsize 25 & \scriptsize 70 & \scriptsize 750 & \scriptsize 99 & \scriptsize150  \\

\scriptsize heart & \scriptsize 14 & \scriptsize 35 & \scriptsize 202 & \scriptsize 27 & \scriptsize40  \\

\scriptsize iris & \scriptsize 5 & \scriptsize 11 & \scriptsize 112 & \scriptsize  15 & \scriptsize22  \\

\hline
\end{tabular}
\end{center}
\label{table:stats}
\end{table}

\begin{table}[h]
\caption{Average test set log likelihoods for structured learning methods on hybrid datasets.}
\label{MSPNvWMISPN-table}
\begin{center}
\begin{tabular}{|p{1.2cm} |p{1cm} p{0.8cm} p{0.8cm} p{0.8cm} p{0.8cm}| }
\hline
\multirow{2}{*}{\bf \small Dataset} &
\multicolumn{1}{c}{\scriptsize WMI-SPN} &
\multicolumn{2}{c}{\scriptsize Gower-MSPN}&
\multicolumn{2}{c|}{\scriptsize RDC-MSPN}\\
& \tiny \ourlearn\ & \quad \tiny hist & \tiny iso & \quad \tiny hist & \tiny iso\\
\hline
\scriptsize anneal-U & \scriptsize \bf -14.543& \scriptsize -63.553 & \scriptsize -38.836 & \scriptsize -60.314 & \scriptsize -38.312 \\

\scriptsize australian & \scriptsize \bf -10.473 & \scriptsize -18.513 & \scriptsize -30.379 & \scriptsize 17.891 & \scriptsize -31.021 \\

\scriptsize auto & \scriptsize \bf -27.126 & \scriptsize -72.998 & \scriptsize -69.405 & \scriptsize -73.378 & \scriptsize -70.066 \\

\scriptsize car & \scriptsize \bf -9.111 & \scriptsize -30.467 & \scriptsize -31.082 & \scriptsize -29.132 & \scriptsize -30.516 \\

\scriptsize cleave & \scriptsize \bf -13.829 & \scriptsize -26.132 & \scriptsize -25.869 & \scriptsize-29.132 & \scriptsize -25.441 \\

\scriptsize crx & \scriptsize \bf -10.525 & \scriptsize -22.422 & \scriptsize -31.624 & \scriptsize -24.036& \scriptsize-31.727 \\

\scriptsize diabetes & \scriptsize \bf -6.299 & \scriptsize -15.286 & \scriptsize -26.968 & \scriptsize -15.930 & \scriptsize-27.242 \\

\scriptsize german & \scriptsize \bf -20.429 & \scriptsize -40.828 & \scriptsize -26.852 & \scriptsize-38.829 & \scriptsize -32.361 \\

\scriptsize german-org & \scriptsize \bf -21.144 & \scriptsize -43.611 & \scriptsize -26.852 & \scriptsize -37.450 & \scriptsize -27.294 \\

\scriptsize heart & \scriptsize \bf -14.875 & \scriptsize -20.691 & \scriptsize -26.994 & \scriptsize -20.376 & \scriptsize -25.906 \\

\scriptsize iris & \scriptsize -3.560 & \scriptsize -3.616 & \scriptsize -2.892 & \scriptsize -3.446 & \scriptsize \bf -2.843 \\

\hline
\end{tabular}
\end{center}
\end{table}

The log-likelihood results for LearnMSPN methods and \ourlearn\  are shown in Table \ref{MSPNvWMISPN-table}. As can be seen, our model outperforms the MSPN model by a wide margin. The performance of our model results in significantly better likelihoods across the majority of datasets. In most cases, our implementation only required two bins per continuous feature, and one instance (the iris dataset)  required a three bin split in order to produce a likelihood competitive to the MSPN models. 

The effectiveness of the framework can be attributed to a number of factors. We observed that unsupervised binning interfaced with LearnSPN's existing structure learning performed well on the hybrid datasets. This can be thought of as piecewise constant WMISPNs. The approach is simple and thus, fast, adding negligible complexity to the original LearnSPN machinery: the effectiveness in learning binary representations and the simple counting of Boolean variable activations is inherited from LearnSPN. In addition to that, we support the learning of polynomial weights, and not just constant or linear ones. In particular, our use of the BIC criteria to choose the most optimal representation, which has bias towards smaller bin numbers and polynomial exponents.

\paragraph{Q2 Model complexity:}
When investigating the correlation of model accuracy to model complexity, we used datasets with the majority variables in the continuous domain. We also used datasets with significantly more instances to learn on, ranges being from 1024 to 17389 in order demonstrate the scalability of WMISPNs. The datasets investigated were the Cloud dataset, Statlog (Shuttle) dataset, and a subset of the MiniBooNE particle identification dataset, which were all from the UCI machine learning repository \citep{Dua:2017}. Our inquires below use equal width binning as a discretisation method for continuous features.

The most immediate question is what is the nature of learned polynomials? The BIC measure, as we mentioned, is used to determine the number of bins and polynomial order. So, by relaxing the criteria for the number of bins to be used, we can study the polynomials. In Table \ref{table:polyprob}, we see statistics on the order of the learned polynomials for 2 vs 5 bins. The order never goes beyond $6$, and in majority of the cases is $\leq 4,$ confirming once again that the learned orders stay manageable. Increasing the bin size favours low-order polynomials: only \emph{australia} and \emph{cloud} have order $6$ polynomials with 5 bins. 

\begin{figure}[h]
  \centering
  \subfloat[Diabetes 2 bin polynomial approximation]{\includegraphics[width=0.4\textwidth]{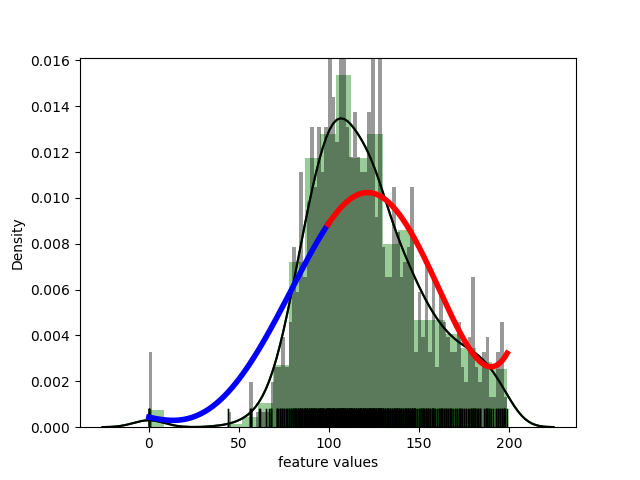}\label{fig:f1d2}}
  \hfill
  \subfloat[Diabetes 5 bin polynomial approximation]{\includegraphics[width =0.4\textwidth]{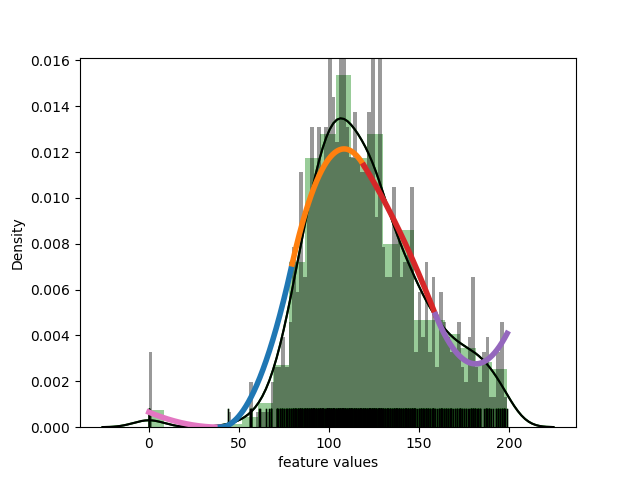}\label{fig:f2d5}}
  \caption{Comparison of learned piecewise polynomial functions for feature 2(Plasma glucose concentration) of the diabetes dataset. Bin intervals are represented by alternating colors.}
  \label{fig:diabpiecewise}
\end{figure}

In Figure \ref{fig:diabpiecewise}, It is evident that increased binning results in piecewise polynomial functions that can better approximate the spread of data. In the case of the diabetes dataset,  at 5 bins the learned polynomial function was better able to capture the distribution of continuous feature points compared with the 2 bin approximation. The improved polynomial function approximation also lends itself to more accurate bin probabilities for SPN inference calculations. Also of note is the polynomial order for the diabetes continuous feature(Plasma glucose concentration) at 2 bins was a $4^{th}$ order polynomial, while at 5 bins the polynomial order was only $2$.

\begin{figure}[h]
\includegraphics[width = 0.45\textwidth]{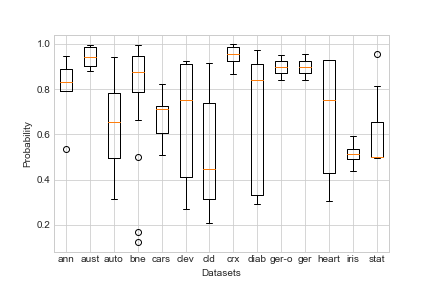}
 \caption{The spread of probabilities, with 2 bins per attribute.}
	\label{fig:spreadQ2}
\end{figure}

The second  question, then, is how are the  probabilities spread from the learned representation? (That is, assume we learn $p_1(x)$ for $0\leq x \leq 5$ and $p_2(x)$ for $5\leq x \leq 10$; we consider the probabilities on computing the volumes and normalising.) Naturally, this spread depends very much on the data, but by considering multiple datasets, one can empirically study the effectiveness of the learning regime. We see in Figure \ref{fig:spreadQ2} that the representations match the characteristics of the data (e.g., spareness for some attributes, missing values), diverse as they are. 

\begin{figure}[h]
\includegraphics[width=0.45\textwidth]{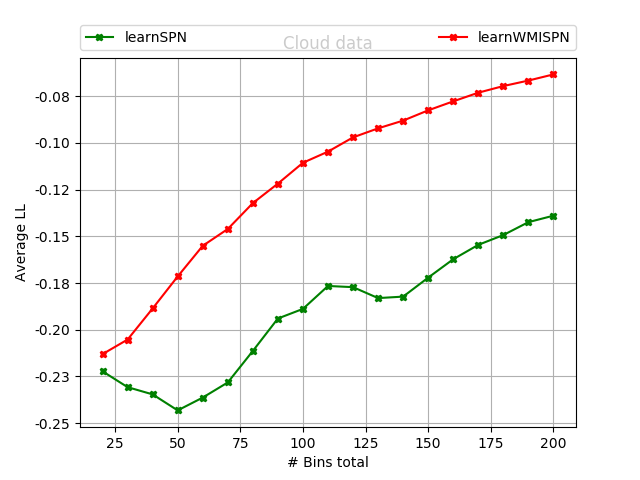}
\caption{Instances of expanded Cloud datasets and resulting normalized log-likelihoods. Comparisons done with \ourlearn\ (red) and LearnSPN (green) \citep{gens2013learning}. It should be noted that \ourlearn\ is yielding a  highly granular and intricate representation. 
For example, at 5 bins per feature, the first bin of the first feature (the pixel mean) defined over a range of $(3.0 \leq x \leq 20.1)$ is given a density approximation that is  a $3^{rd}$ degree polynomial, and then at 15 bins per feature, over a range of $(3.0 \leq x \leq 8.7)$ the density approximation is now a $4^{th}$ degree polynomial.} 
\label{fig:curveQ2}
\end{figure}

The third natural question is whether learning polynomials are beneficial at all? We mentioned earlier that unsupervised binning along with a simple binarisation scheme can be seen as a simplistic hybrid model -- an instance of piecewise constant WMISPNs -- for which LearnSPN suffices. Clearly such an endeavour would come at a significant loss of expressiveness, e.g., no interval querie. So, by letting BIC determine the polynomial order but explicitly setting the bin parameter, we can contrast LearnSPN and \ourlearn. It should be noted that as each bin range corresponds to a distribution represented by a given polynomial, further bin increases on a feature result in differing polynomial representations. (Analogously, classical SPNs will redistribute the spread of discrete probabilities.) In Figure \ref{fig:curveQ2}, the plotted performance of average log-likelihoods over bin complexity is normalized by the number of new variables generated from equal width binning. We see that accuracy on the dataset does increase as the number of bins per feature increases, and thus \ourlearn\ yields a more accurate representation.

\begin{table}[h]
\centering{ \footnotesize 
\begin{tabular}{|p{1.2cm} |p{0.3cm} p{0.7cm} p{0.7cm} p{0.7cm} p{0.7cm} p{0.7cm}|} 
\hline 
\small \textbf{Dataset}    & \scriptsize \textbf{Bins} & \scriptsize \textbf{2nd-Order} & \scriptsize \textbf{3rd-Order} & \scriptsize \textbf{4th-Order} & \scriptsize \textbf{5th-Order} & \scriptsize \textbf{6th-Order}  \\
\hline 

\scriptsize australia  & \scriptsize 2    & \scriptsize 0         & \scriptsize 16.667    & \scriptsize 16.667    & \scriptsize 33.3     & \scriptsize 33.3      \\
\scriptsize australia  & \scriptsize 5    & \scriptsize 16.667    & \scriptsize 33.333    & \scriptsize 16.667    & \scriptsize 16.667    & \scriptsize 16.667     \\ \hline
\scriptsize auto       & \scriptsize 2    & \scriptsize 15.8     & \scriptsize 36.842    & \scriptsize 31.579    & \scriptsize  12.789    & \scriptsize 0          \\
\scriptsize auto       & \scriptsize 5    & \scriptsize 73.684    & \scriptsize 15.789    & \scriptsize 10.526    & \scriptsize 0         & \scriptsize 0          \\ \hline
\scriptsize german-org & \scriptsize 2    & \scriptsize 0         & \scriptsize 33.333    & \scriptsize 0         & \scriptsize 0         & \scriptsize 66.666     \\
\scriptsize german-org & \scriptsize 5    & \scriptsize 33.333    & \scriptsize 33.333    & \scriptsize 0         &  \scriptsize 33.333    & \scriptsize 0          \\ \hline
\scriptsize heart      & \scriptsize 2    & \scriptsize 0         & \scriptsize 20        & \scriptsize 20        & \scriptsize 60        & \scriptsize 0          \\
\scriptsize heart      & \scriptsize 5    & \scriptsize 0         & \scriptsize 60        & \scriptsize 40        & \scriptsize 0         & \scriptsize 0          \\ \hline
\scriptsize iris       & \scriptsize 2    & \scriptsize 50        & \scriptsize 0         & \scriptsize 50        & \scriptsize 0         & \scriptsize 0          \\
\scriptsize iris       & \scriptsize 5    & \scriptsize 75        & \scriptsize 25        & \scriptsize 0         & \scriptsize 0         & \scriptsize 0        \\ \hline
\scriptsize statlog    & \scriptsize 2    & \scriptsize 42.857    & \scriptsize 0         & \scriptsize 0         & \scriptsize 0         & \scriptsize 57.143     \\ 
\scriptsize statlog    & \scriptsize 5    & \scriptsize 28.571    & \scriptsize 57.143    & \scriptsize 0         & \scriptsize 14.289    & \scriptsize 0          \\\hline
\scriptsize cloud      & \scriptsize 2    & \scriptsize 10        & \scriptsize 10        & \scriptsize 40        & \scriptsize 10        & \scriptsize 30         \\
\scriptsize cloud      & \scriptsize 5    & \scriptsize 30        & \scriptsize 50        & \scriptsize 10        & \scriptsize 0         & \scriptsize 10         \\\hline  
\end{tabular}
\caption{Comparison of the distributions of orders (in \%) for 2 and 5 bins.}\label{table:polyprob}}

\end{table}

\paragraph{Q3 Query Interface:}
The capacity to query SPNs trained on continuous domains represents a significant improvement in terms of interpreting data. 
With regard to posing complex queries to WMISPNs, we studied the computation time for inferences. The three datasets used for measuring this are the Cloud, Statlog (Shuttle), and MiniBooNE datasets. All datasets are comprised of continuous features, with Statlog containing a single discrete feature. In order to measure complexity, we define query length of $i$ as length$(Q_i) = q_i$ with query length of 1 corresponding to a query of form $Q_1 =(a \leq x < b)$ in the continuous case and $Q_1=(\neg y)$ in the discrete case. It follows that a query length of two $q_2$ could be defined as $Q_2 = [(a \leq x < b) \wedge (\neg y)]$ and so forth. Continuous queries are defined as $c^i$ and discrete as $d^i$ with $i$ representing the query count. We generated 10 random queries based on the continuous ranges for each dataset, including discrete outcomes for Statlog, and averaged the inference time. We also fixed the number of bins per continuous feature to two.

\begin{table}[h]
\caption{Average query time for differing query lengths. Time per query is measured in (ns/query).}
\label{queryspeed-table}
\begin{center}
\begin{tabular}{|p{1.0cm} |p{0.3cm} |p{0.35cm} p{0.35cm} |p{0.35cm} p{0.35cm} |p{0.35cm} p{0.35cm}| p{0.35cm} p{0.35cm}|}
\hline
\multirow{1}{*}{\bf \small Dataset} &
\multicolumn{1}{c|}{\footnotesize $q_1$} &
\multicolumn{2}{c|}{\footnotesize $q_2$}&
\multicolumn{2}{c|}{\footnotesize $q_3$}&
\multicolumn{2}{c|}{\footnotesize $q_4$}&
\multicolumn{2}{c|}{\footnotesize $q_5$}\\
& \tiny $c^1$ &  \tiny $c^2$ & \tiny $c^1,d^1$& \tiny $c^3$ & \tiny $c^2,d^1$ & \tiny $c^4$ & \tiny $c^3,d^1$ & \tiny $c^5$ & \tiny $c^4,d^1$\\
\hline
\scriptsize cloud & \scriptsize 1401& \scriptsize 2154 & \scriptsize n/a & \scriptsize 2563& \scriptsize n/a & \scriptsize 3659& \scriptsize n/a & \scriptsize 4025 & \scriptsize n/a\\

\scriptsize statlog & \scriptsize 1299 & \scriptsize n/a & \scriptsize 1878 & \scriptsize n/a & \scriptsize 2365 & \scriptsize n/a & \scriptsize 2749 & \scriptsize n/a & \scriptsize 2982\\

\scriptsize 
miniboone & \scriptsize 1168 & \scriptsize 2071 & \scriptsize n/a & \scriptsize 2219 & \scriptsize n/a & \scriptsize 2290 & \scriptsize n/a & \scriptsize 3048 & \scriptsize n/a\\

\hline
\end{tabular}
\end{center}
\end{table}

In Table \ref{queryspeed-table}, we see the inference speeds on the complex queries. Overall, it is clear that complex queries do not slow down the model. For all query lengths, the time per query remains in nanosecond range. As query length increases, more time is required but overall the increase in query time is linear. The model was also capable of handling queries with mixed continuous and discrete atoms, further demonstrating the capability of WMISPNs.  

\section{Conclusion}
Deep architectures are powerful learning paradigms that capture latent structure and have proven to be very successful in machine learning. Guessing the right architecture for complex data is challenging, and so paradigms such as SPNs are attractive alternatives in providing an  \textit{unsupervised} learning regime in addition to  robust inference computations.

Here, we pushed the envelope further to consider a systematic integration of SPNs and WMI, allowing us to learn tractable, non-parametric distributions and convex combinations thereof for hybrid data, also in an \textit{unsupervised} fashion. The integration was achieved by minimally adapting the base case for a SPN structure learning module, which makes our approach generic to a large extent. Different from our predecessors, we show for the first time how tractable distributions of arbitrary granularity can be learned, and more importantly, how to query these distributions over a rich interval syntax. Our empirical results show that our implemented system is effective, scalable and incurs very little cost for handling continuous features, all of which is very desirable for learning from big uncertain data. Challenges for the future include extending the query language with features like counting operators, which would allow us to reason about the cardinality of sets of objects in an image, thus enabling an interface for commonsensical reasoning within deep architectures. 

\section{Acknowledgements}

This work is partly supported by the EPSRC grant `Towards Explainable and Robust Statistical AI: A Symbolic Approach'. 
\bibliography{main}

\begin{thebibliography}{30}
\providecommand{\natexlab}[1]{#1}
\providecommand{\url}[1]{\texttt{#1}}
\expandafter\ifx\csname urlstyle\endcsname\relax
  \providecommand{\doi}[1]{doi: #1}\else
  \providecommand{\doi}{doi: \begingroup \urlstyle{rm}\Url}\fi

\bibitem[Albarghouthi et~al.(2017)Albarghouthi, D'Antoni, Drews, and
  Nori]{albarghouthi2017quantifying}
A.~Albarghouthi, L.~D'Antoni, S.~Drews, and A.~Nori.
\newblock Quantifying program bias.
\newblock \emph{arXiv preprint arXiv:1702.05437}, 2017.

\bibitem[Bacchus et~al.(2009)Bacchus, Dalmao, and
  Pitassi]{DBLP:journals/jair/BacchusDP09}
F.~Bacchus, S.~Dalmao, and T.~Pitassi.
\newblock Solving {\#}{SAT} and {B}ayesian inference with backtracking search.
\newblock \emph{J. Artif. Intell. Res. {(JAIR)}}, 34:\penalty0 391--442, 2009.

\bibitem[Bach and Jordan(2002)]{bach2002thin}
F.~R. Bach and M.~I. Jordan.
\newblock Thin junction trees.
\newblock In \emph{Advances in Neural Information Processing Systems}, pages
  569--576, 2002.

\bibitem[Baldoni et~al.(2014)Baldoni, Berline, De~Loera, Dutra, K{\"o}ppe,
  Moreinis, Pinto, Vergne, and Wu]{baldoni2014user}
V.~Baldoni, N.~Berline, J.~A. De~Loera, B.~Dutra, M.~K{\"o}ppe, S.~Moreinis,
  G.~Pinto, M.~Vergne, and J.~Wu.
\newblock A user's guide for latte integrale v1. 7.1.
\newblock \emph{Optimization}, 22:\penalty0 2, 2014.

\bibitem[Bekker et~al.(2015)Bekker, Davis, Choi, Darwiche, and Van~den
  Broeck]{bekker2015tractable}
J.~Bekker, J.~Davis, A.~Choi, A.~Darwiche, and G.~Van~den Broeck.
\newblock Tractable learning for complex probability queries.
\newblock In \emph{Advances in Neural Information Processing Systems}, pages
  2242--2250, 2015.

\bibitem[Belle et~al.(2015)Belle, Passerini, and Van~den
  Broeck]{belle2015probabilistic}
V.~Belle, A.~Passerini, and G.~Van~den Broeck.
\newblock Probabilistic inference in hybrid domains by weighted model
  integration.
\newblock In \emph{Proceedings of 24th International Joint Conference on
  Artificial Intelligence (IJCAI)}, pages 2770--2776, 2015.

\bibitem[Bengio(2009)]{bengio2009learning}
Y.~Bengio.
\newblock Learning deep architectures for {AI}.
\newblock \emph{Foundations and trends in Machine Learning}, 2\penalty0
  (1):\penalty0 1--127, 2009.

\bibitem[Chavira and Darwiche(2008)]{chavira2008probabilistic}
M.~Chavira and A.~Darwiche.
\newblock On probabilistic inference by weighted model counting.
\newblock \emph{Artificial Intelligence}, 172\penalty0 (6-7):\penalty0
  772--799, 2008.

\bibitem[Chistikov et~al.(2015)Chistikov, Dimitrova, and Majumdar]{RupakSMT}
D.~Chistikov, R.~Dimitrova, and R.~Majumdar.
\newblock Approximate counting in smt and value estimation for probabilistic
  programs.
\newblock In \emph{TACAS}, volume 9035, pages 320--334. ACTA INFORMATICA, 2015.

\bibitem[Dheeru and Karra~Taniskidou(2017)]{Dua:2017}
D.~Dheeru and E.~Karra~Taniskidou.
\newblock {UCI} machine learning repository, 2017.
\newblock URL \url{http://archive.ics.uci.edu/ml}.

\bibitem[Fierens et~al.(2011)Fierens, Van~den Broeck, Thon, Gutmann, and
  Raedt]{DBLP:conf/uai/FierensBTGR11}
D.~Fierens, G.~Van~den Broeck, I.~Thon, B.~Gutmann, and L.~D. Raedt.
\newblock Inference in probabilistic logic programs using weighted {CNF}'s.
\newblock In \emph{UAI}, pages 211--220, 2011.

\bibitem[Gens and Domingos(2013)]{gens2013learning}
R.~Gens and P.~Domingos.
\newblock Learning the structure of sum-product networks.
\newblock In \emph{International Conference on Machine Learning}, pages
  873--880, 2013.

\bibitem[Heckerman et~al.(1995)Heckerman, Geiger, and Chickering]{HGC95}
D.~Heckerman, D.~Geiger, and D.~Chickering.
\newblock Learning {Bayesian} networks: The combination of knowledge and
  statistical data.
\newblock \emph{Machine Learning}, 20:\penalty0 197--243, 1995.

\bibitem[Hsu et~al.(2017)Hsu, Kalra, and Poupart]{hsu2017online}
W.~Hsu, A.~Kalra, and P.~Poupart.
\newblock Online structure learning for sum-product networks with gaussian
  leaves.
\newblock \emph{arXiv preprint arXiv:1701.05265}, 2017.

\bibitem[Lauritzen and Jensen(2001)]{MixtureofGaussians}
S.~L. Lauritzen and F.~Jensen.
\newblock Stable local computation with conditional gaussian distributions.
\newblock \emph{Statistics and Computing}, 11\penalty0 (2):\penalty0 191--203,
  2001.

\bibitem[Liang et~al.(2017)Liang, Bekker, and Van~den
  Broeck]{liang2017learning}
Y.~Liang, J.~Bekker, and G.~Van~den Broeck.
\newblock Learning the structure of probabilistic sentential decision diagrams.
\newblock In \emph{Proceedings of the 33rd Conference on Uncertainty in
  Artificial Intelligence (UAI)}, 2017.

\bibitem[Molina et~al.(2017)Molina, Natarajan, and Kersting]{molina2017poisson}
A.~Molina, S.~Natarajan, and K.~Kersting.
\newblock Poisson sum-product networks: A deep architecture for tractable
  multivariate poisson distributions.
\newblock In \emph{AAAI}, pages 2357--2363, 2017.

\bibitem[Molina et~al.(2018)Molina, Vergari, Di~Mauro, Natarajan, Esposito, and
  Kersting]{molina2018mixed}
A.~Molina, A.~Vergari, N.~Di~Mauro, S.~Natarajan, F.~Esposito, and K.~Kersting.
\newblock Mixed sum-product networks: A deep architecture for hybrid domains.
\newblock In \emph{AAAI}, 2018.

\bibitem[Morettin et~al.(2017)Morettin, Passerini, Sebastiani,
  et~al.]{morettin2017efficient}
P.~Morettin, A.~Passerini, R.~Sebastiani, et~al.
\newblock Efficient weighted model integration via smt-based predicate
  abstraction.
\newblock In \emph{IJCAI}, pages 720--728, 2017.

\bibitem[Murphy(1999)]{murphy1999variational}
K.~P. Murphy.
\newblock A variational approximation for bayesian networks with discrete and
  continuous latent variables.
\newblock In \emph{UAI}, pages 457--466, 1999.

\bibitem[Nitti et~al.(2016)Nitti, Ravkic, Davis, and
  De~Raedt]{nitti2016learning}
D.~Nitti, I.~Ravkic, J.~Davis, and L.~De~Raedt.
\newblock Learning the structure of dynamic hybrid relational models.
\newblock In \emph{22nd European Conference on Artificial Intelligence (ECAI)
  2016}, volume 285, pages 1283--1290, 2016.

\bibitem[Poon and Domingos(2011)]{poon2011sum}
H.~Poon and P.~Domingos.
\newblock Sum-product networks: A new deep architecture.
\newblock \emph{Proc. 12th Conf. on Uncertainty in Artificial Intelligence},
  pages 337--346, 2011.

\bibitem[Ravkic et~al.(2015)Ravkic, Ramon, and Davis]{ravkic2015learning}
I.~Ravkic, J.~Ramon, and J.~Davis.
\newblock Learning relational dependency networks in hybrid domains.
\newblock \emph{Machine Learning}, 100\penalty0 (2-3):\penalty0 217--254, 2015.

\bibitem[Sanner and Abbasnejad(2012)]{sanner2012symbolic}
S.~Sanner and E.~Abbasnejad.
\newblock Symbolic variable elimination for discrete and continuous graphical
  models.
\newblock In \emph{AAAI}, 2012.

\bibitem[Schwarz(1978)]{schwarz1978estimating}
G.~Schwarz.
\newblock Estimating the dimension of a model.
\newblock \emph{The annals of statistics}, 6\penalty0 (2):\penalty0 461--464,
  1978.

\bibitem[Shenoy and West(2011)]{shenoy2011extended}
P.~P. Shenoy and J.~C. West.
\newblock Extended shenoy--shafer architecture for inference in hybrid bayesian
  networks with deterministic conditionals.
\newblock \emph{International Journal of Approximate Reasoning}, 52\penalty0
  (6):\penalty0 805--818, 2011.

\bibitem[Speichert(2017)]{Speichert:Thesis:2017}
S.~Speichert.
\newblock {Learning Hybrid Relational Rules with Piecewise Polynomial Weight
  Functions for Probabilistic Logic Programming}.
\newblock Master's thesis, The University of Edinburgh, Scotland, 2017.

\bibitem[Valiant(1979)]{valiant1979complexity}
L.~G. Valiant.
\newblock The complexity of enumeration and reliability problems.
\newblock \emph{SIAM Journal on Computing}, 8\penalty0 (3):\penalty0 410--421,
  1979.

\bibitem[Vergari et~al.(2015)Vergari, Di~Mauro, and
  Esposito]{vergari2015simplifying}
A.~Vergari, N.~Di~Mauro, and F.~Esposito.
\newblock Simplifying, regularizing and strengthening sum-product network
  structure learning.
\newblock In \emph{Joint European Conference on Machine Learning and Knowledge
  Discovery in Databases}, pages 343--358. Springer, 2015.

\bibitem[Yang et~al.(2014)Yang, Cao, and Yu]{yang2014exponential}
X.~Yang, J.~Cao, and W.~Yu.
\newblock Exponential synchronization of memristive cohen--grossberg neural
  networks with mixed delays.
\newblock \emph{Cognitive neurodynamics}, 8\penalty0 (3):\penalty0 239--249,
  2014.

\end{thebibliography}
\end{document}